\documentclass{article}

\usepackage[final]{neurips_2019}

\usepackage[utf8]{inputenc} 
\usepackage[T1]{fontenc}    
\usepackage{hyperref}       
\usepackage{url}            
\usepackage{booktabs}       
\usepackage{amsfonts}       
\usepackage{nicefrac}       
\usepackage{microtype}      
\usepackage{amsmath}
\usepackage{amssymb}
\usepackage{xcolor}
\usepackage{multirow,multicol,graphicx}
\usepackage{subcaption}
\usepackage{makecell}
\usepackage{textcomp}

\renewcommand{\textuparrow}{$\uparrow$}
\renewcommand{\textdownarrow}{$\downarrow$}

\title{Out-of-distribution Detection in Classifiers via Generation}

\author{
Sachin Vernekar\textsuperscript{1}\\
\texttt{sverneka@uwaterloo.ca}\\
\And
Ashish Gaurav\textsuperscript{1}\\
\texttt{a5gaurav@uwaterloo.ca}\\
\And
Vahdat Abdelzad\textsuperscript{2}\\
\texttt{vabdelza@gsd.uwaterloo.ca}\\
\And
Taylor Denouden\textsuperscript{1}\\
\texttt{tadenoud@uwaterloo.ca}\\
\And
Rick Salay\textsuperscript{2}\\
\texttt{rsalay@gsd.uwaterloo.ca}\\
\And
Krzysztof Czarnecki\textsuperscript{2}\\
\texttt{kczarnec@gsd.uwaterloo.ca}\\
\\
Department of Computer Science\textsuperscript{1}\\
Department of Electrical and Computer Engineering \textsuperscript{2}\\
University of Waterloo\\
}
\begin{document}

\maketitle

\begin{abstract}

By design, discriminatively trained neural network classifiers produce reliable predictions only for in-distribution samples. For their real-world deployments, detecting out-of-distribution (OOD) samples is essential. Assuming OOD to be outside the closed boundary of in-distribution, typical neural classifiers do not contain the knowledge of this boundary for OOD detection during inference. There have been recent approaches to instill this knowledge in classifiers by explicitly training the classifier with OOD samples close to the in-distribution boundary. However, these generated samples fail to cover the entire in-distribution boundary effectively, thereby resulting in a sub-optimal OOD detector. In this paper, we analyze the feasibility of such approaches by investigating the complexity of producing such ``effective'' OOD samples. We also propose a novel algorithm to generate such samples using a manifold learning network (e.g., variational-autoencoder) and then train an n+1 classifier for OOD detection, where the $n+1^{th}$ class represents the OOD samples. We compare our approach against several recent classifier-based OOD detectors on MNIST and Fashion-MNIST datasets. Overall the proposed approach consistently performs better than the others.

\end{abstract}

\section{Introduction}

A typical neural classifier is trained on in-distribution data belonging to a fixed set of known classes. The classifier makes reliable predictions when the test input belongs to in-distribution. However, in many real-world deployments of these models (e.g., object detection in autonomous driving), the test input could potentially be sampled from any part of the input space, including the region where the support of the in-distribution is negligible. The predictions of the classifier on such samples are arbitrary. Indeed, recent results on image classification models have shown that the deep neural network (DNN) classifiers make overconfident predictions (\cite{overConfidence}) not just on randomly generated OOD images but also on semantically meaningful images belonging to OOD (\cite{natrualAdversaries}). This necessitates the use of OOD detectors in applications where input data is not guaranteed to be sampled from the in-distribution.

\begin{figure}[t]
    \centering
    \begin{subfigure}[t]{.5\textwidth}
        \centering
        \includegraphics[width=0.8\columnwidth]{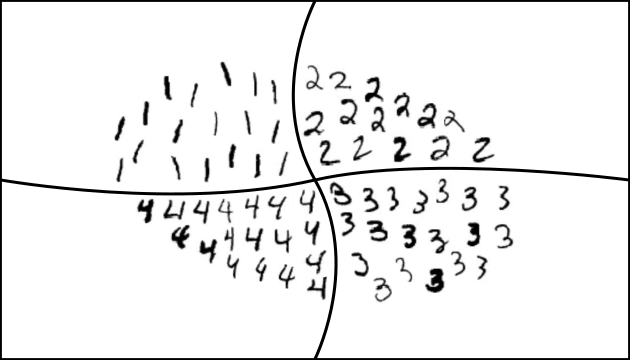}
        \caption{}
        \label{fig:1a}
    \end{subfigure}\hfill
    \begin{subfigure}[t]{.5\textwidth}
        \centering
        \includegraphics[width=0.8\columnwidth]{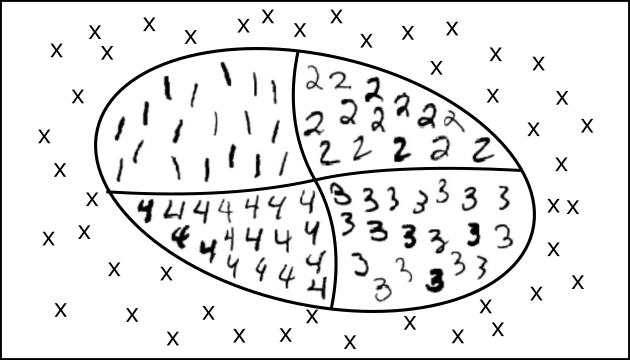}
        \caption{}
        \label{fig:1b}
    \end{subfigure}
    \caption{Decision boundaries change and become more bounded around in-distribution when a classifier is trained with an $n+1^{th}$ class containing OOD samples. (a) Unbounded decision boundaries of a typical 4-class classifier.  (b) A 5-class classifier trained with outlier samples ‘x’ forming the fifth class, that are close to in-distribution resulting in bounded decision boundaries around in-distribution.}
    \label{fig:1}
\end{figure}

Decision boundary of a typical $n$-class softmax classifier divides the input space into a set of known classes. However, these regions extend beyond the support of in-distribution leading to over-generalization as shown in Figure \ref{fig:1a}.  \cite{lee2018training} attempt to address this problem by explicitly training a classifier with generated out-of-distribution samples that follow the low-density boundary of in-distribution in addition to the in-distribution samples. They show that for effective OOD detection, the generated OOD samples should cover and be close to the low-density boundary of in-distribution. The approach uses a ``confidence-loss’’ to train a  ``confident-classifier,’’ where the standard cross-entropy is minimized (minimize output entropy) over in-distribution data. For the generated out-of-distribution samples, a KL loss is minimized to make the classifier outputs follow a uniform distribution (maximize output entropy). The assumption here is that the effect of maximizing the entropy for OOD samples close to the in-distribution propagates to regions far away from in-distribution where the support of in-distribution is negligible. Therefore, if we construct a decision boundary based on the output entropy, the predictions in the in-distribution regions will have lower entropy and higher entropy for anything outside the in-distribution regions. Thus, we could detect OOD samples based on the output entropy. The OOD data is generated by a joint training of a GAN and a classifier.  \cite{sricharan2018building} also follow a similar approach where the objectives of GAN and classifier are reversed w.r.t. the OOD data. 

\cite{vernekar2019analysis} analyze the confident-classifier and show that the assumption that maximizing the output entropy for OOD data in the low-density boundary of in-distribution results in higher entropy for points far away from the in-distribution is unrealistic and instead suggest to train an n+1 class classifier where the $n+1^{th}$ class represents the generated OOD samples. They empirically show that the resulting decision boundaries of the $n+1$ class classifier would divide the input space into regions such that the space containing in-distribution data are classified into one of the first n classes, and the rest of the input space becomes the OOD class ($n+1^{th}$ class). We follow this approach in this paper in dealing with the generated OOD samples.

\textbf{Contribution.}
In this work, we propose a novel algorithm for generating ``effective'' OOD samples. The approach leverages the following generic assumptions (\cite{manifoldHypothesis1}, \cite{manifoldHypothesis2}, \cite{tangentClassifier}) that holds true for a wide range of problems, primarily for image data, which is the data used to validate our approach.

\textbf{The manifold hypothesis} states that the higher dimensional real-world data in the input space is likely concentrated on a much lower-dimensional sub-manifold.

\textbf{The multi-class manifold hypothesis} states that, if the data contains multiple classes, different classes correspond to disjoint sub-manifolds separated by low-density regions in the input space.

\begin{figure}[t]
    \centering
    \begin{subfigure}[t]{.5\textwidth}
        \centering
        \frame{\includegraphics[width=0.8\columnwidth, height=1.5in]{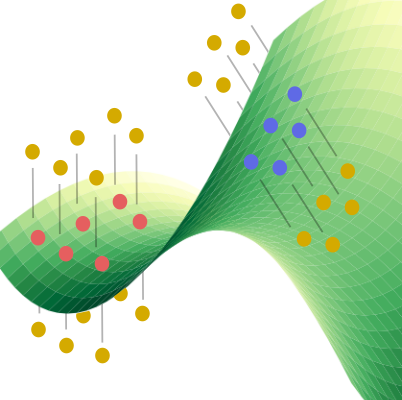}}
        \caption{Type I}
    \end{subfigure}\hfill
    \begin{subfigure}[t]{.5\textwidth}
        \centering
        \frame{\includegraphics[width=0.8\columnwidth, height=1.5in]{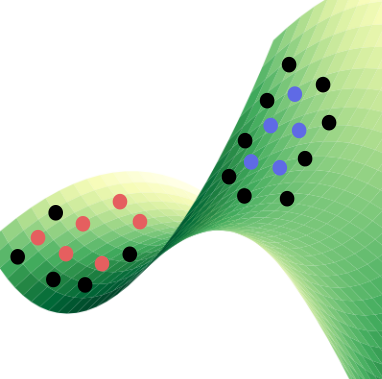}}
        \caption{Type II}
    \end{subfigure}
    \caption{Categories of outliers that we generate: (a) Type I (yellow), which includes samples that are close to the data but outside the in-distribution sub-manifolds, and (b) Type II (black), which includes samples that lie on the in-distribution sub-manifolds and trace the in-distribution boundary; in-distribution clusters are represented through blue and red points.}
    \label{fig:2}
\end{figure}

To fully cover the ``boundary'' of in-distribution, we identify two categories of OOD samples that are to be generated. As shown in Figure \ref{fig:2}, \textbf{Type I)} are the OOD samples that are close but outside the in-distribution sub-manifolds; \textbf{Type II)} are the OOD samples that are on the sub-manifolds but close to the ``boundary‘’ of the in-distribution.

The assumption that the auto-encoders in general model the lower dimensional data-supporting manifold is well established in the literature. Therefore we use a conditional variational auto-encoder (CVAE) for generating ``effective'' OOD samples. These generated OOD samples are then used to train an $n+1$ class classifier as suggested in \cite{vernekar2019analysis}. In an ideal case, if these samples cover the entire boundary of in-distribution, the resulting decision boundary of the classifier would look like the one shown in Figure \ref{fig:1b}, where the region in the input space where the support of in-distribution is negligible is classified as OOD class, and the rest is mapped to the inlier classes.

The type I OOD samples are generated by adding an epsilon perturbation to input samples in the direction perpendicular to the tangent space of the sub-manifold at that point. These directions are obtained as follows. We take the Jacobian of the decoder of CVAE, whose column vectors span the tangent space of the sub-manifold at that point. The basis vectors of the left-nullspace of the Jacobian span the space perpendicular to the sub-manifold at that point. Type II OOD samples are obtained by sampling from the low-density regions of the aggregate approximate posterior of the CVAE. These samples in latent space are then decoded to obtain the OOD samples needed.

We compare the performance of our OOD detector against a suite of classifiers-based OOD detectors on MNIST and Fashion-MNIST datasets as in-distributions. Overall, our approach does better than most of the out-of-distribution datasets. Especially, when compared against the confident-classifier baseline that also relies on training with generated OOD samples, our approach outperforms this approach by a substantial margin in most of our experiments. This can be taken as an evidence to the fact that, the OOD samples generated by our approach are more diverse and cover the in-distribution boundaries better compared to the ones generated by GAN in the confident-classifier approach. This is also visually apparent from the experimental results on a 3D-dataset shown in section \ref{OOD sample generation} of appendix. Moreover, we argue that our method also gives an intuitive measure of the complexity of generating such ``effective'' OOD samples.

\section{Related Work}

There have been many approaches in the literature proposed to address the problem of OOD detection in the context of image data. Most of the  successful ones are either generative (\citep{aaeOOD, saferClassification, likelihoodRatio})  or classifier-based approaches \citep{hendrycks2016baseline, hendrycks2018deep, devries2018learning, liang2018enhancing, simpleUnifiedApproach}. Generative approaches either explicitly or implicitly estimate the input density or use reconstruction error as a criterion to decide if input belongs to OOD. Classifier-based approaches, on the other hand, incorporate OOD detection as a part of the classifier network. The approach proposed in this paper belongs to the latter category. Therefore we limit our related work discussion to only classifier-based approaches.

Typical discriminatively trained classifiers that model the conditional probability $P(y|x)$ without any additional constraints, by definition can make reliable classification decisions only on in-distribution data. For out-of-distribution data, the classifier output is arbitrary. Moreover, any meta information from the output of the classifier or the features learned are also conditioned on the data belonging to in-distribution. Therefore this information in-principle cannot be used to ascertain if the input is in or out of distribution. However, most of the recent approaches in the literature follow this approach.

\cite{hendrycks2016baseline} propose a baseline approach
to detect OOD inputs, called max-softmax by thresholding the maximum softmax output of a pre-trained classifier. \cite{liang2018enhancing} improve upon this using temperature scaling (ODIN, \citep{guo2017calibration}) and adding input perturbations. The assumptions is that these changes result in larger separation between in and out of distribution data in terms of their output predictions. \cite{simpleUnifiedApproach} propose an approach based on the assumption that the class-conditional features of a softmax classifier follow a Gaussian distribution. Therefore, Mahalanobis distance (MD) from the mean of the Gaussian is used as a score to detect OOD. This is then combined with input perturbations similar to ODIN to enhance the OOD detection results. This method obtains state-of-the-art results on most of the baseline datasets used in OOD detection literature.  Despite good results, the method can be seen as OOD detection on feature space rather than pixel space not conforming to the usual definition of OOD (By definition, the in-distribution, $p_{in}(x)$ is defined for $x \in \mathbb{X}$ in pixel space, and hence OOD is also defined in the same space). Hence the effectiveness of the method highly depends on the features learned by the classifier, and also there is no guarantee that the optimization algorithm forces the features to follow a Gaussian distribution. \cite{hendrycks2018deep} propose to train a classifier with a confidence loss where OOD data is sampled from a large natural dataset. \cite{hein2019relu} also follow a similar approach using a confidence loss and uniformly generated random OOD samples from the input space. In addition, they not only minimize the confidence at the generated OOD samples, but also in the neighbourhood of those samples. However, because both these approaches use the confidence-loss, they suffer from the same problems highlighted in \cite{vernekar2019analysis}. Moreover, such approaches are only feasible for input spaces where it is possible to represent the support of OOD with finite samples (assuming uniform distribution over OOD). This is not possible when the input space is $\mathbb{R}^d$, whereas our approach is still applicable.

\cite{geifman2018boosting} propose to use Bayesian prediction uncertainties given by MC-Dropout \citep{gal2016dropout} for OOD detection. However, on the theoretical front, the Bayesian uncertainty measure only characterizes the uncertainty in in-distribution. Therefore in principle should not be applied to OOD detection.

\section{Out-of-distribution Sample Generation}

For effective OOD detection, the generated samples should effectively cover the entire in-distribution ``boundary’’. From the manifold hypothesis, the data density concentrates near a low dimensional manifold in a high dimensional space. Therefore as mentioned earlier, we divide generated OOD samples into two categories.

\subsection{OOD samples outside the data manifold}

These samples are obtained by adding small perturbations to in-distribution samples that are concentrated on the manifold.\ These perturbations should be added in directions such that the resulting samples should fall outside the manifold. The directions locally normal to the data-supporting manifold can be thought of as the directions that are less likely to contain in-distribution samples and the tangent directions as the more likely ones. Therefore we add perturbations in the normal directions to get OOD samples.

Deep generative models such as VAEs \citep{kingma2013auto} and GANs \citep{goodfellow2014generative} can model the data manifold of observations $x \in X$ through corresponding latent variables $z \in Z$ via a mapping function $g:Z \rightarrow X$ as $x=g(z)$. With a choice of reasonably lower dimensional $z$ and a flexible generative function $g$, the model can efficiently represent the true data manifold. Following the \textbf{multi-class manifold hypothesis}, we use a conditional generative model that is conditioned over the class labels. For our experiments, we use a conditional variational auto-encoder although the choice between Conditional VAEs and GANs is arbitrary. 

Let $h: X \rightarrow Z$ and $g: Z \rightarrow \hat{X}$ denote the encoder and decoder functions of CVAE, respectively. The tangent space of the manifold at a point $x \in X$ is given by the column space of the Jacobian\footnote{While $z$ is stochastic, we just use its mean estimate for generating OOD samples outside the manifold.}

\begin{equation}
J(x) = \frac{\partial g(z)}{\partial z}\bigg|_{z=h(x)} 
\end{equation}

\begin{figure}[t]
    \centering
    \begin{subfigure}[t]{.245\textwidth}
        \centering
        \includegraphics[width=\columnwidth]{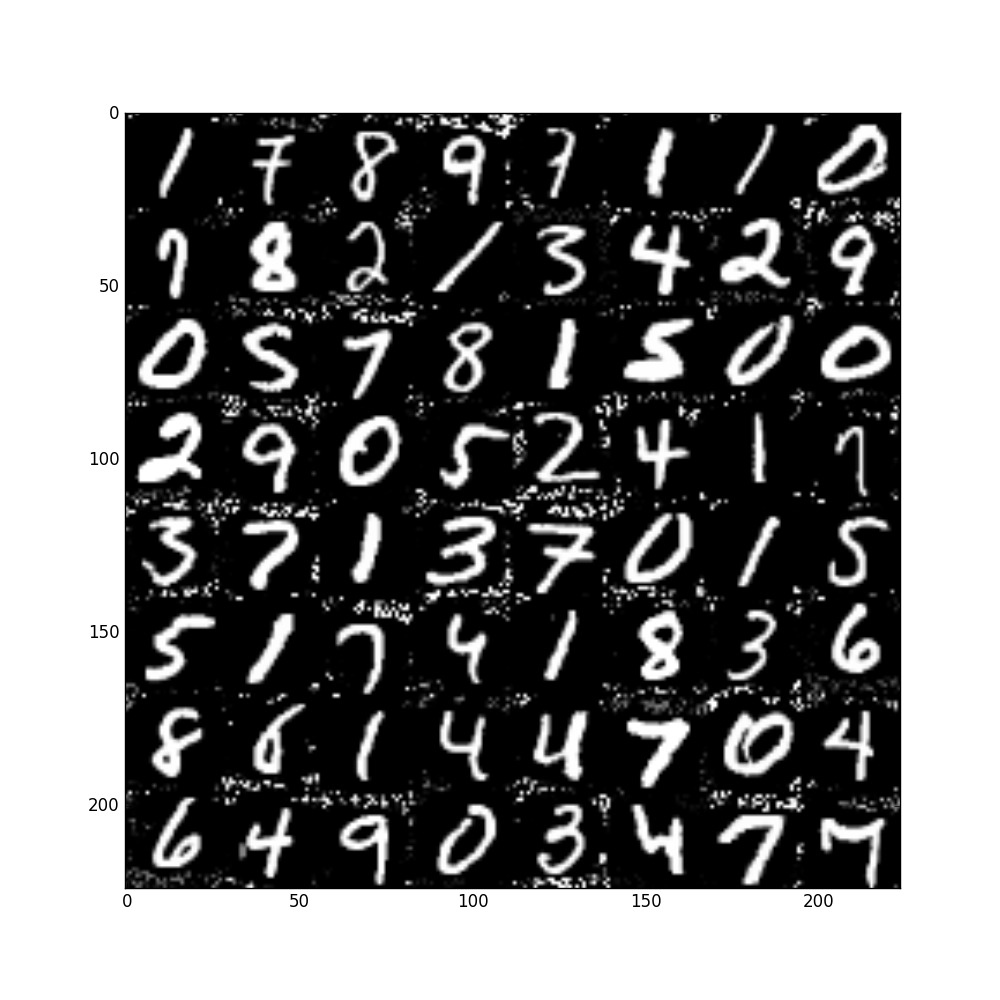}
        \caption{MNIST (Type I)}
    \end{subfigure}
    \begin{subfigure}[t]{.245\textwidth}
        \centering
        \includegraphics[width=\columnwidth]{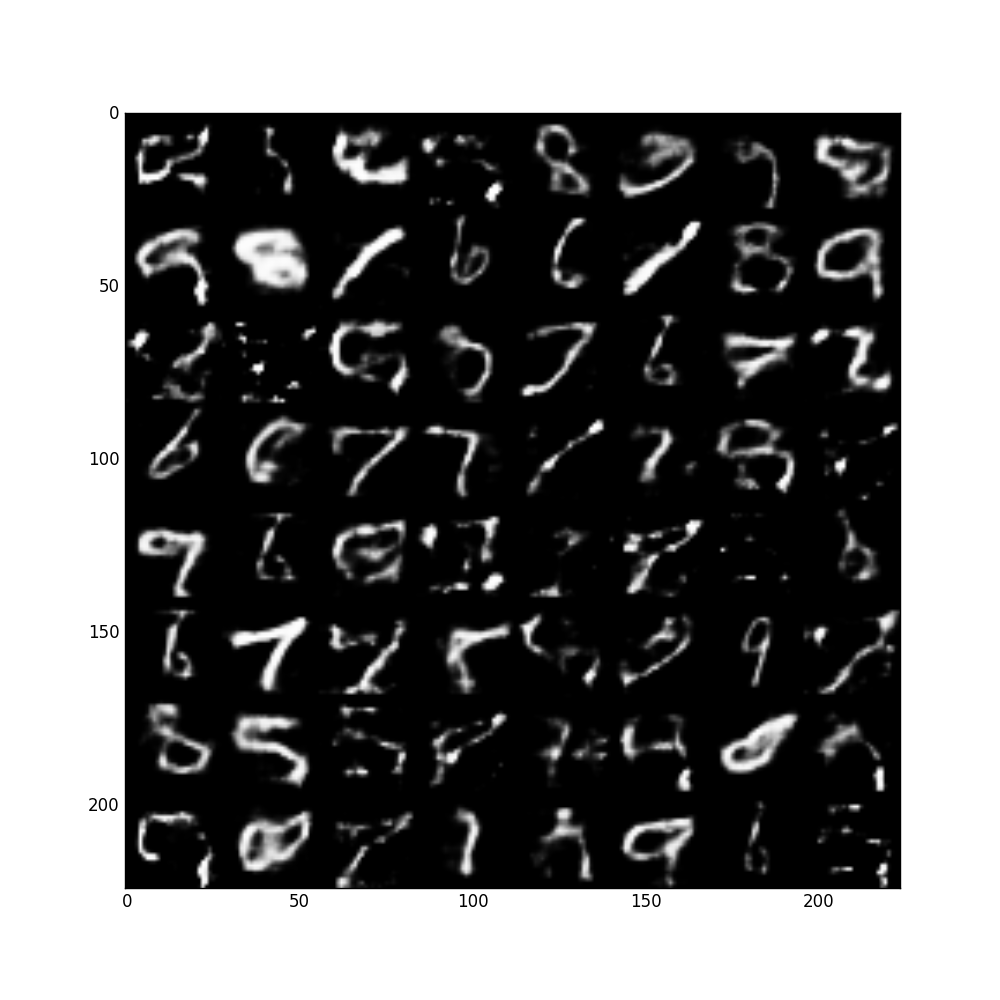}
        \caption{MNIST (Type II)}
    \end{subfigure}
    \begin{subfigure}[t]{.245\textwidth}
        \centering
        \includegraphics[width=\columnwidth]{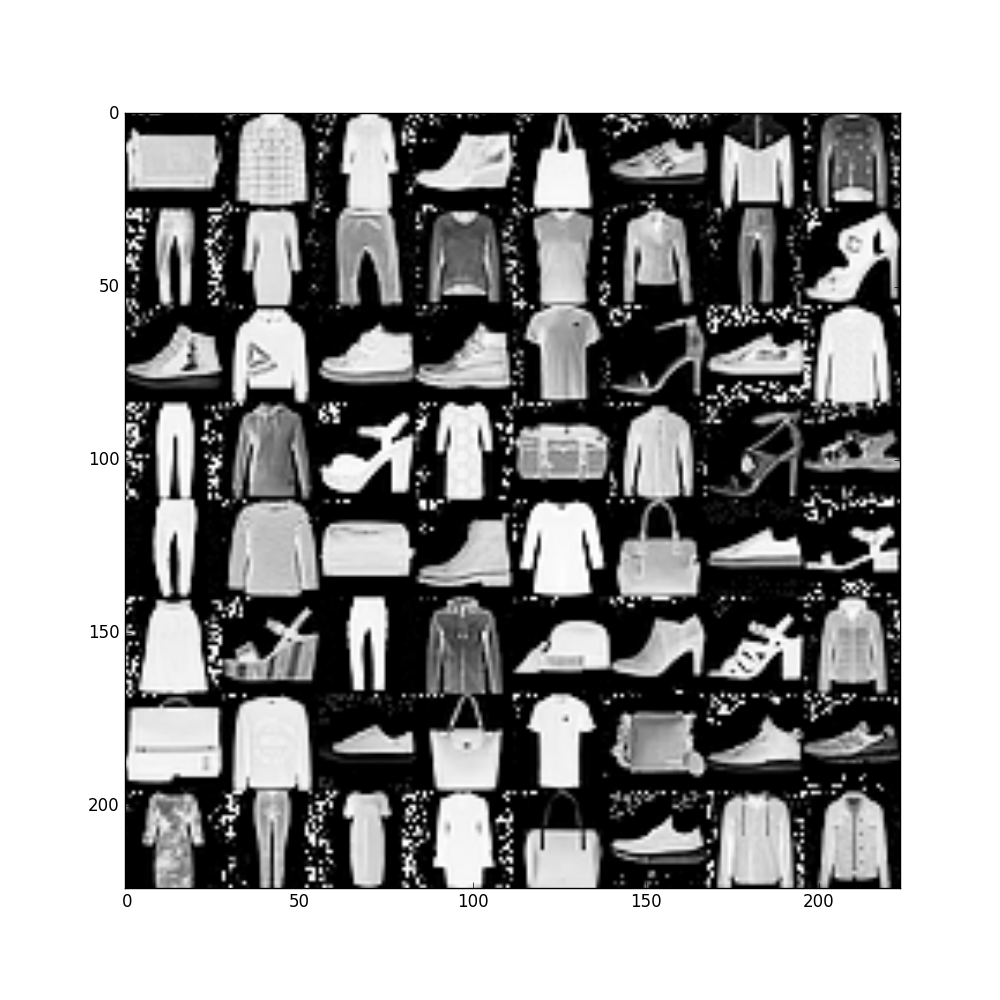}
        \caption{F-MNIST (Type I)}
    \end{subfigure}
    \begin{subfigure}[t]{.245\textwidth}
        \centering
        \includegraphics[width=\columnwidth]{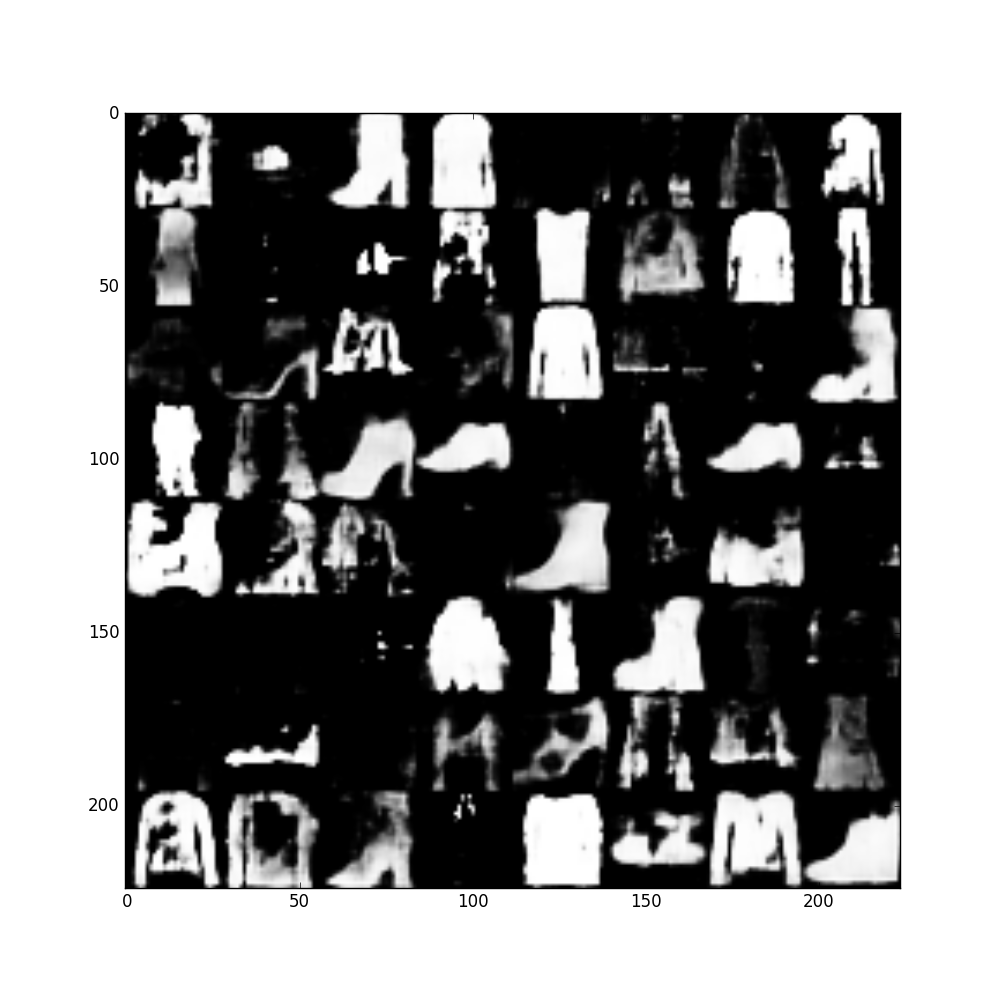}
        \caption{F-MNIST (Type II)}
    \end{subfigure}
    \caption{Generated outlier samples using the proposed method; Type I outliers typically modify the background pixels (normal components have the least variance), while Type II outliers modify the object pixels.}
    \label{fig:3}
\end{figure}

Let $N(x)$ denote the null-space of $J^T(x)$ (left null space of $J(x)$). Then the basis vectors of $N(x)$ span the normal bundle of the manifold at $x$. Let $v(x)\sim N(x)$ be a randomly sampled unit vector from $N(x)$, then the perturbed sample is given by,

\begin{align}
    \Tilde{x} = x + \beta v(x)
\end{align}

where $\beta \in \mathbb{R}$ is a hyper-parameter that controls how far the perturbed sample is from the in-distribution point. In our experiments, we use a stochastic $\beta$ that is uniformly sampled from in the range $[0.1, 1.0]$. Figure \ref{fig:3} illustrates the perturbed samples for MNIST and Fashion MNIST datasets. One can observe that the perturbations added mostly modify the background pixels than the object pixels. This is because the normal directions to the manifold mostly represent least variance components of the image. The GAN based OOD generation (\citep{lee2018training}), as indicated in \cite{vernekar2019analysis}, fails to produce diverse OOD samples to effectively cover the in-distribution boundary. However, our approach generates OOD samples by randomly perturbing in-distribution training samples, hence the diversity of the generated samples is ensured.

\subsection{OOD samples on the data manifold}

These are the samples that are in the low-density regions of the input space but close to the in-distribution boundaries on the manifold. 

For a variational auto-encoder, the aggregate posterior $q(z)$ \citep{adversarialAutoencoders} is given by, 

\begin{align}
q(z) &= \int_{x} q(z|x) p_{in}(x) dx
\end{align} 

where $p_{in}(x)$ is the probability density function of in-distribution and $q(z/x)$ is the approximate posterior. Assuming a smooth decoder, the high-density regions in the aggregate posterior can be thought of as corresponding to densely populated regions in the input space, and the input space density would gradually decrease as we sample away from the high-density regions in the aggregate posterior. Therefore the in-distribution boundary on the manifold can be approximated by regions at a distance away from the high-density areas where the density dips below a certain threshold. We approximate $q(z)$ with a uni-modal Gaussian distribution whose mean  $\hat{\mu}$ and co-variance $\hat{\Sigma}$ are estimated using the encoder mappings of in-distribution samples. We use Mahalanobis distance as a criterion to determine the distance from the mean to sample and generate the required OOD samples. Let $r$ be the Mahalanobis distance from the mean of $q(z)$ that encompasses 95\% of the training data. The OOD samples are generated by decoding the uniformly sampled samples from the latent space over the surface of a hyper-ellipsoid \citep{hyperEllipsoid} defined by \ref{ellipsoid}, where $\hat{\mu}_z$ and $\hat{\Sigma}_z$ are the mean and co-variance estimates of $q(z)$, respectively.

\begin{align}
(z-\hat{\mu}_{z})^T \hat{\Sigma}^{-1}_{z} (z-\hat{\mu}_{z}) = r^2\label{ellipsoid}
\end{align}

It is fair to assume a uni-modal Gaussian distribution for $q(z)$ as we fit a Gaussian per class. Moreover, a substantial gain in the ODD detection results when the classifier is trained with these samples can also be taken as evidence pointing towards the validity of such an assumption. 

The generated OOD samples are then used to train an $n+1$ class softmax classifier, where the $n+1^{th}$ class represents the OOD class. Regular cross-entropy loss is used to train the classifier as suggested in \cite{vernekar2019analysis}. The OOD class prediction probability as the OOD score.

\section{Experiments}

We validated\footnote{code is available at \href{https://github.com/sverneka/OODGen}{https://github.com/sverneka/OODGen}} our approach on MNIST and Fashion MNIST as in-distribution datasets and several other OOD datasets. For all MNIST as in-distribution experiments, we use a CVAE with a latent dimension of 8, and for Fashion MNIST, the latent dimension is set to 10. We compare our approach against the recent classifier-based OOD detectors such as confident-classifier, ODIN and Mahalanobis distance-based approach without feature ensemble (MD). The architecture for both CVAE and the classifier used are shown in the appendix. Both the networks are trained till convergence.

\subsection{OOD Datasets}

MNIST is used as an OOD dataset for Fashion MNIST \citep{xiao2017/online} as in-distribution, and vice-versa. For MNIST 0-4 experiment, we use images in class 0 through 4 as in-distribution and class 5 through 9 as OOD. The other datasets used as OOD for all our experiments, including the synthetic ones, are listed below. 

\textbf{Omniglot} \citep{Lake1332} contains different handwritten characters from 50 different alphabets. The images are downsampled to $28\times28$.

\textbf{EMNIST-letters} \citep{cohen2017emnist} contains hand-written English alphabets. This is one of the challenging datasets for MNIST as in-distribution experiments given its similarity of MNIST as both are hand-written characters. Therefore, most of the baselines tend to perform worse on this compared to other OOD datasets. 

\textbf{NotMNIST} \citep{bulatov2011notmnist} is similar to MNIST, except that it contains synthetic images of characters A through I of various fonts.

\textbf{Gaussian noise} includes gray-scale images, where each pixel is sampled from an independent normal distribution with 0.5 mean and unit-variance.

\textbf{Uniform noise} includes gray-scale images where each pixel is sampled from an independent uniform distribution in the range $[0, 1]$.

\textbf{Sphere OOD} contains images sampled from the surface of a 784 dimensional hyper-sphere centered at the origin with a radius equal to the maximum Euclidean distance of in-distribution samples from the origin and reshaped to $28\times28$. This is used to show the effectiveness of our approach not only on the datasets that are restricted to a finite range such as $[0, 1]^d$ for images in $[0, 1]^d$ but also for a general case of $\mathbb{R}^d$.

\subsection{Evaluation metrics for OOD detection}
We experimented with two different metrics as OOD score to determine if the given input sample is in or out of distribution. \textbf{OOD class probability} is the $n+1^{th}$ class prediction probability. \textbf{In-distribution max probability} is  the maximum prediction probabilities of the in-distribution classes. A higher (lower) OOD class probability (in-distribution max probability) indicates a higher probability of a sample being OOD. Except for MNIST 0-4 experiments, we find that the former metric gives the best results. We report only the best score in \ref{results-table}. We use the area under the ROC curve (AUROC\textuparrow ), the area under the precision-recall curve (AUPRC\textuparrow), the false positive rate at 95\% true positive rate (FPR95\textdownarrow) and the detection error as the metrics for evaluation. These metrics are commonly used for evaluating OOD detection methods \citep{hendrycks2016baseline, lee2018training, hendrycks2018deep}.

\subsection{Detection Results}
Table 1 compares our approach with other approaches for experiments on MNIST and Fashion MNIST as in-distribution datasets. Since the classifier is trained with OOD samples, there is a possibility of reduction in the classification accuracy of in-distribution classes. We therefore report classification accuracy of a classifier trained with and without OOD samples. We find that there is no significant change in accuracy. Training our method requires tuning hyper-parameter such as $\beta$ from Eq. 2, OOD class weight, and learning rate. For all our experiments we use a stochastic $\beta$ uniformly sampled in the range [0.1, 1], OOD class weight is set to 0.1 and Adadelta \citep{zeiler2012adadelta} is the optimizer used with learning rates of 0.1 and 0.01 for Fashion-MNIST and MNIST experiments, respectively. The hyper-parameters were chosen based on the in-distribution classification accuracy and the AUROC with NotMNIST as the OOD dataset for MNIST experiments and EMNIST as the OOD dataset for Fashion MNIST experiments. We don't tune the hyper parameters per OOD dataset unlike ODIN and Mahalanobis distance-based approaches, where the perturbation magnitude is tuned per OOD dataset. Even without this advantage, our method still performs better than these baselines for most of the OOD datasets.

We would like to remark that our approach gives good OOD detection results consistently on all the OOD datasets used unlike the baselines compared. This indicates that our approach is robust to change in OOD datasets.

\begin{table}
    \scriptsize
  \caption{Results}
  \label{results-table}
  \hskip-1.5cm\begin{tabular}{lllllll}
    \toprule
    \multirow{2}{*}{\makecell{ID Model\\(acc before OOD/\\acc after OOD)}}     & \multirow{2}{*}{OOD}    & \makecell{FPR at \\ 95\% TPR \textdownarrow} & \makecell{Detection \\ Error\textdownarrow} & \makecell{AUROC\textuparrow} & \makecell{AUPR\\Out\textuparrow} & \makecell{AUPR\\In\textuparrow} \\
    \cmidrule(r){3-7}
    & & \multicolumn{5}{c}{Ours/Confident-Classifier/ODIN/MD} \\
    \midrule
    \multirow{7}{*}{\makecell{MNIST\\(99.0/98.9)}} & F-MNIST  & \textbf{0.0}/7.9/0.4/94.2 & \textbf{0.2}/5.6/1.8/11.9 & \textbf{100.0}/98.5/99.8/86.6 & \textbf{100.0}/98.8/99.8/92.0 & \textbf{100.0}/98.4/99.8/74.0    \\
    & EMNIST-letters & \textbf{1.6}/31.0/25.7/31.2 & \textbf{3.0}/13.2/11.7/13.6 & \textbf{99.6}/93.0/94.4/93.2 & \textbf{99.6}/93.0/94.3/92.7 & \textbf{99.6}/92.4/94.1/93.2   \\
    & NotMNIST & \textbf{0.0}/26.5/11.3/34.8 & \textbf{0.0}/12.3/6.9/16.3 & \textbf{100.0}/94.0/97.8/91.7 & \textbf{100.0}/93.9/\textbf{100.0}/91.7 & \textbf{100.0}/93.8/97.7/92.3 \\
    & Omniglot & \textbf{0.0}/\textbf{0.0}/\textbf{0.0}/98.5 & \textbf{0.0}/1.0/0.2/46.9 & \textbf{100.0}/\textbf{100.0}/\textbf{100.0}/19.8 & \textbf{100.0}/\textbf{100.0}/\textbf{100.0}/40.8 & \textbf{100.0}/\textbf{100.0}/\textbf{100.0}/35.0
    \\
    & Gaussian-Noise & \textbf{0.0}/\textbf{0.0}/\textbf{0.0}/99.9 & \textbf{0.0}/\textbf{0.0}/\textbf{0.0}/24.6 & \textbf{100.0}/\textbf{100.0}/\textbf{100.0}/50.9 & \textbf{100.0}/\textbf{100.0}/\textbf{100.0}/71.8 & \textbf{100.0}/\textbf{100.0}/\textbf{100.0}/35.1
    \\
    & Uniform-Noise & \textbf{0.0}/\textbf{0.0}/\textbf{0.0}/82.6 & \textbf{0.0}/\textbf{0.0}/\textbf{0.0}/26.4 & \textbf{100.0}/\textbf{100.0}/\textbf{100.0}/65.0 & \textbf{100.0}/\textbf{100.0}/\textbf{100.0}/76.0 & \textbf{100.0}/\textbf{100.0}/\textbf{100.0}/63.9
    \\
    & Sphere-OOD & \textbf{0.0}/21.6/0.0/80.4 & \textbf{0.1}/6.6/1.4/14.9 & \textbf{100.0}/96.8/99.8/87.6 & \textbf{100.0}/97.8/99.9/91.7 & \textbf{100.0}/95.2/99.8/79.9 \\
    \midrule
    \multirow{7}{*}{\makecell{F-MNIST\\(91.9/91.2)}} & MNIST  & 4.1/87.4/70.2/\textbf{2.4} & 4.2/36.3/28.9/\textbf{3.6} & 98.7/67.0/76.7/\textbf{99.5} & 98.2/65.2/73.2/\textbf{99.5} & \textbf{100.0}/64.8/77.3/99.4    \\
    & EMNIST-letters & \textbf{6.4}/87.3/83.5/10.1 & \textbf{5.4}/41.8/13.6/7.3 & 97.9/61.1/66.6/\textbf{98.1} & 96.8/60.0/62.0/\textbf{98.3} & \textbf{98.5}/61.6/66.6/98.1   \\
    & NotMNIST & \textbf{0.8}/78.9/80.2/7.2 & \textbf{1.2}/32.2/33.9/5.8 & \textbf{99.7}/73.7/69.3/97.8 & \textbf{99.5}/73.0/63.0/97.4 & \textbf{99.8}/72.4/70.5/98.2 \\
    & Omniglot & \textbf{0.0}/59.8/9.6/58.4 & \textbf{0.9}/22.1/7.1/26.8 & \textbf{99.8}/85.6/97.9/83.2 & \textbf{99.9}/85.8/97.6/84.9 & \textbf{99.6}/85.1/98.2/83.4
    \\
    & Gaussian-Noise & \textbf{0.0}/32.2/4.5/99.7 & \textbf{0.2}/9.6/3.8/19.9 & \textbf{99.8}/95.8/98.0/80.0 & \textbf{99.9}/96.7/96.7/87.0 & \textbf{99.5}/94.7/95.6/66.3
    \\
    & Uniform-Noise & \textbf{0.2}/71.0/99.4/1.7 & \textbf{1.3}/16.4/24.7/3.3 & \textbf{99.8}/88.6/74.7/98.9 & \textbf{99.8}/91.8/82.9/99.2 & \textbf{99.8}/82.9/61.6/97.9
    \\
    & Sphere-OOD & 0.6/99.3/100.0/\textbf{0.0} & 0.8/50.0/50.0/\textbf{0.0} & 99.7/29.6/0.25/\textbf{100.0} & 99.4/39.1/30.7/\textbf{100.0} & 99.8/37.4/30.7/\textbf{100.0} \\
    \midrule
    \multirow{7}{*}{\makecell{MNIST0-4\\(99.8/99.5)}} & MNIST5-9  & 23.0/21.9/\textbf{20.4}/50.0 & 12.1/12.0/\textbf{11.5}/14.4 & \textbf{93.8}/92.9/93.4/92.3 & 92.5/92.1/91.3/\textbf{93.8} & \textbf{94.2}/93.6/\textbf{94.2}/90.1    \\
    & F-MNIST & \textbf{0.4}/1.7/2.0/41.4 & \textbf{1.6}/3.1/3.4/15.1 & \textbf{99.9}/99.4/99.4/92.5 & \textbf{99.9}/99.5/99.4/93.3 & \textbf{99.9}/99.3/99.3/91.9   \\
    & EMNIST-letters & \textbf{3.7}/22.1/26.4/12.9 & \textbf{4.2}/12.4/13.9/7.6 & \textbf{99.1}/92.9/92.3/96.9 & \textbf{99.1}/92.0/90.4/96.6 & \textbf{99.1}/93.6/93.2/97.1   \\
    & NotMNIST & \textbf{0.0}/10.9/28.0/2.8 & \textbf{0.1}/7.7/13.3/3.1 & \textbf{100.0}/97.5/93.5/99.3 & \textbf{100.0}/97.5/92.7/99.2 & \textbf{100.0}/97.6/93.7/99.4 \\
    & Omniglot & \textbf{0.0}/\textbf{0.0}/2.3/\textbf{0.0} & \textbf{0.0}/0.1/3.6/0.4 & \textbf{100.0}/\textbf{100.0}/99.1/\textbf{100.0} & \textbf{100.0}/\textbf{100.0}/99.3/\textbf{100.0} & \textbf{100.0}/\textbf{100.0}/98.8/\textbf{100.0}
    \\
    & Gaussian-Noise & \textbf{0.0}/\textbf{0.0}/\textbf{0.0}/0.2 & \textbf{0.0}/\textbf{0.0}/0.1/2.4 & \textbf{100.0}/\textbf{100.0}/\textbf{100.0}/97.5 & \textbf{100.0}/\textbf{100.0}/\textbf{100.0}/98.6 & \textbf{100.0}/\textbf{100.0}/99.7/92.2
    \\
    & Uniform-Noise & \textbf{0.0}/\textbf{0.0}/\textbf{0.0}/25.9 & \textbf{0.0}/\textbf{0.0}/0.4/5.1 & \textbf{100.0}/\textbf{100.0}/99.9/95.9 & \textbf{100.0}/\textbf{100.0}/99.9/97.6 & \textbf{100.0}/\textbf{100.0}/99.6/89.4
    \\
    & Sphere-OOD & 0.5/7.1/\textbf{0.2}/22.7 & \textbf{1.3}/5.5/2.0/6.9 & \textbf{99.9}/98.2/99.6/96.5 & \textbf{99.9}/98.6/99.7/97.6 & \textbf{99.9}/97.4/99.3/93.8 \\
    \midrule
    \multirow{7}{*}{\makecell{F-MNIST0-4\\(94.2/94.8)}} & F-MNIST5-9  & \textbf{19.7}/55.8/29.2/75.8 & \textbf{12.3}/17.1/14.6/26.4 & \textbf{92.5}/89.5/92.1/79.5 & 88.7/90.2/\textbf{91.3}/79.8 & \textbf{94.3}/87.1/92.8/77.7    \\
    & MNIST & \textbf{1.8}/67.3/53.5/2.0 & \textbf{2.3}/23.6/21.1/3.4 & \textbf{99.5}/83.5/86.4/99.0 & \textbf{99.4}/84.2/86.1/99.3 & \textbf{99.6}/81.7/85.7/98.5   \\
    & EMNIST-letters & \textbf{1.2}/71.6/48.4/14.1 & \textbf{2.4}/24.2/20.3/7.6 & \textbf{99.6}/82.6/87.9/97.6 & \textbf{99.6}/83.8/87.7/98.0 & \textbf{99.7}/79.8/87.9/96.9 \\
    & NotMNIST & \textbf{0.2}/76.0/57.7/11.0 & \textbf{1.2}/26.8/23.6/8.0 & \textbf{99.9}/79.9/84.1/97.0 & \textbf{99.8}/81.3/83.8/96.8 & \textbf{99.9}/77.1/83.9/97.2
    \\
    & Omniglot & \textbf{1.0}/62.3/15.5/11.1 & \textbf{2.5}/18.3/9.1/7.1 & \textbf{99.5}/88.6/96.5/97.5 & \textbf{99.6}/90.6/96.0/97.9 & \textbf{99.3}/85.8/96.7/95.7
    \\
    & Gaussian-Noise & \textbf{0.0}/0.3/\textbf{0.0}/99.3 & \textbf{0.4}/2.0/\textbf{0.4}/41.7 & \textbf{100.0}/99.7/\textbf{100.0}/53.4 & \textbf{100.0}/99.8/\textbf{100.0}/62.6 & \textbf{100.0}/99.7/\textbf{100.0}/47.9
    \\
    & Uniform-Noise & \textbf{0.0}/9.8/1.3/36.3 & \textbf{0.3}/5.4/3.0/8.5 & \textbf{100.0}/98.1/99.2/95.0 & \textbf{100.0}/98.6/99.4/96.6 & \textbf{100.0}/97.5/98.9/90.1
    \\
    & Sphere-OOD & \textbf{0.0}/89.6/95.5/\textbf{0.0} & \textbf{0.0}/38.3/41.6/\textbf{0.0} & \textbf{100.0}/65.8/59.8/\textbf{100.0} & \textbf{100.0}/67.7/62.4/\textbf{100.0} & \textbf{100.0}/61.9/55.0/\textbf{100.0} \\
    \bottomrule
  \end{tabular}
\end{table}

\section{Discussion \& Conclusion}
We propose a novel algorithm for generating ``effective'' OOD samples for training an $n+1$-class classifier for OOD detection. For generating OOD samples outside the manifold, we randomly sample from the left-null-space of the Jacobian as described earlier. But the complexity of this step depends on the number of basis vectors in the null-space and its dimensions. For the MNIST case, with the input dimensions $28\times28$ and latent dimension of 8, there are 776 basis vectors in the left-nullspace. To cover the in-distribution boundary effectively in all directions, many OOD samples for each in-distribution training sample are to be generated by taking random linear combination of the basis vectors, which is quite expensive. This gives an intuitive measure of effective OOD sample complexity. However, we find that only a few OOD samples are sufficient to guide the decision boundary of the classifier to be bounded around the in-distribution regions evidenced by their OOD detection results.

While the proposed approach gets good results on gray-scale datasets, in future revisions we would like to investigate its effectiveness on non-gray-scale datasets, such as CIFAR and ImageNet.

\bibliography{mybib.bib}
\bibliographystyle{apalike}

\appendix

\section{Generating OOD samples using a GAN vs Our approach} \label{OOD sample generation}

\cite{lee2018training} propose a joint training of GAN and a classifier based on the following objective:
\begin{align}
        \min_G \max_D \min_\theta &\underbrace{\mathbb{E}_{P_{in}(\hat{x},\hat{y})}[-\log P_\theta (y=\hat{y}|\hat{x})]}_{\text{(a)}} + \beta \underbrace{\mathbb{E}_{P_{G}(x)}[\text{KL}(\mathcal{U}(y)|| P_\theta(y|x))]}_\text{(b)} \nonumber \\
        &+ \underbrace{\mathbb{E}_{P_{in}(x)}[\log D(x)] + \mathbb{E}_{P_G(x)}[\log(1-D(x))]\label{(1)}}_\text{(c)}
\end{align}
where (b)+(c) is the modified GAN loss and (a)+(b) is the classifier loss ($\theta$ is the classifier's parameter) called the confidence loss. The difference from the regular GAN objective is the additional KL loss in (\ref{(1)}), which when combined with the original loss, forces the generator to generate samples in the low-density boundaries of the in-distribution ($P_{in}(x)$) space. $\beta$ is a hyper-parameter that controls how close the OOD samples are to the in-distribution boundary. For the classifier, the KL loss pushes the OOD samples generated by GAN to produce a uniform distribution at the output, and therefore have higher entropy. This enables one to detect OOD samples based on the entropy or the confidence at the output of the classifier.

With a toy experiment, they show that the generator indeed produces such samples and also these samples follow the ``boundary'' of the in-distribution data. However, in the experiment, they use a pre-trained classifier. The classifier is pre-trained to optimize the confidence loss on in-distribution and OOD samples sampled close to the in-distribution. Therefore the classifier already has the knowledge of those OOD samples. When GAN is then trained following the objective in (\ref{(1)}), GAN likely generates those OOD samples close to the in-distribution. But it is evident that this setting is not realistic as one cannot have a fully informative prior knowledge of those OOD samples if our objective is to generate them.

\begin{figure}[t]
    \centering
    \begin{subfigure}[t]{.5\textwidth}
        \centering
        \includegraphics[width=0.8\columnwidth, height=1.5in]{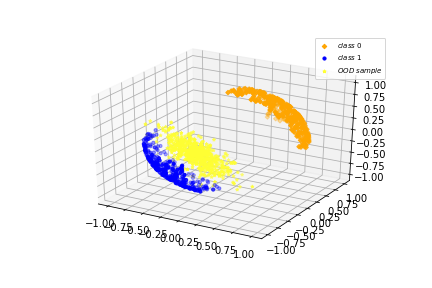}
        \caption{3d plot}
    \end{subfigure}\hfill
    \begin{subfigure}[t]{.5\textwidth}
        \centering
        \includegraphics[width=0.8\columnwidth, height=1.5in]{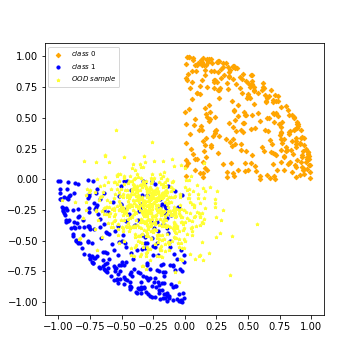}
        \caption{2d projection}
    \end{subfigure}
    \caption{Generated OOD samples using a joint training of a GAN and a confident-classifier. We observe that the generated OOD samples don't cover the entire in-distribution boundary.}
    \label{fig:exp3}
\end{figure}

Therefore, we experiment by directly optimizing (\ref{(1)}) where the classifier is not pre-trained. The in-distribution data for the experiment is obtained by sampling over the surface of a unit sphere from its diagonally opposite quadrants to form 2 classes respectively as shown in Figure \ref{fig:exp3}. We find that (with much hyper-parameter tuning), even though GAN ends up producing OOD samples close to the in-distribution, it does an unsatisfactory job at producing samples that could follow the entire in-distribution boundary. Moreover, there is less diversity in the generated samples which make them ineffective at improving the classifier performance in OOD detection. Our intuition is that the loss (\ref{(1)}(b)+\ref{(1)}(c)) that forces the generator of the GAN to generate samples in the high entropy regions of the classifier doesn't necessarily enforce it to produce samples that follow the entire in-distribution boundary. The inability of GANs to generate such samples for a simple 3D dataset indicates that it would be even more difficult in higher dimensions.

\begin{figure}[!htb]
\subfigure{0.32\textwidth}
  \includegraphics[width=\linewidth]{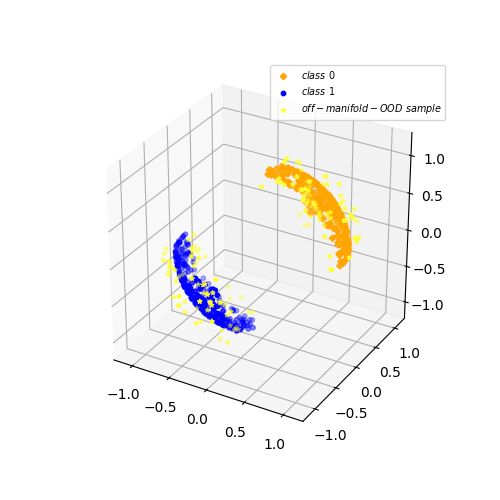}
  \caption{}\label{fig:exp3_new_a}
\endsubfigure\hfill
\subfigure{0.32\textwidth}
  \includegraphics[width=\linewidth]{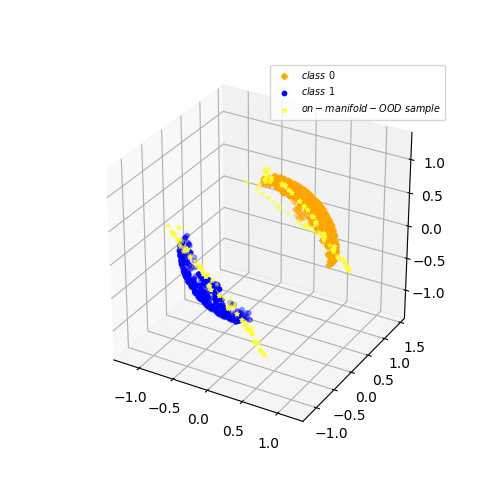}
  \caption{}\label{fig:exp3_new_b}
\endsubfigure\hfill
\subfigure{0.32\textwidth}%
  \includegraphics[width=\linewidth]{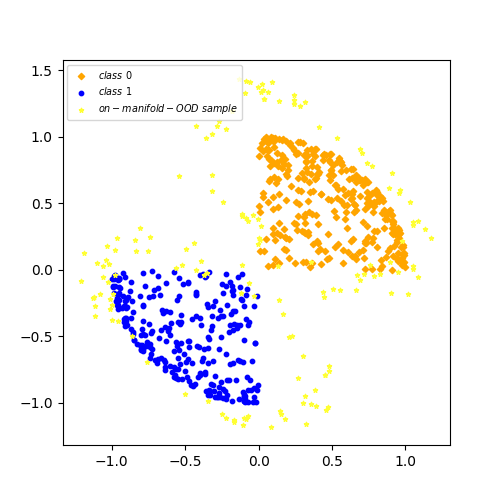}
  \caption{}\label{fig:exp3_new_c}
\endsubfigure
\caption{Generated boundary OOD samples using our approach. (a) 3d plot of in-distribution data with out-of-manifold boundary OOD samples. (b) 3d plot of in-distribution data with on-manifold boundary OOD samples. (c) 2d projection of in-distribution data with on-manifold boundary samples to show that they cover the in-distribution boundary on the manifold.}
\label{fig:exp3_new}
\end{figure}

In comparison to the GAN based boundary OOD generation, our approach as visually apparent from Figure \ref{fig:exp3_new} produces samples that cover the in-distribution boundary quite effectively. While it is difficult to visualize how well the off-manifold OOD samples cover the boundary, one can imagine them having a good coverage on the off-manifold boundary as they are obtained by perturbing each training sample in the direction given by the null-spaces. Hence the diversity of the OOD samples is ensured. For on-manifold boundary OOD samples, as evident from Figure \ref{fig:exp3_new_c}, it forms a closed boundary around the in-distribution points.

\section{Experimental Architecture}

\begin{figure}[t]
    \centering
    \begin{subfigure}[t]{0.9\textwidth}
        \centering
        \includegraphics[width=\columnwidth]{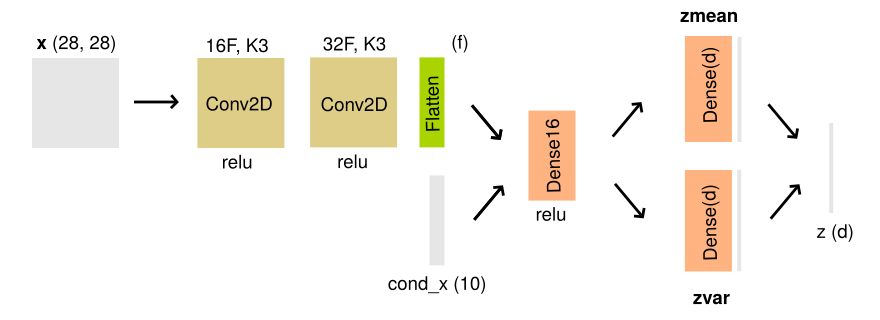}
        \caption{Encoder architecture}
        \label{fig:architecture-enc}
    \end{subfigure}
    \begin{subfigure}[t]{\textwidth}
        \centering
        \includegraphics[width=\columnwidth]{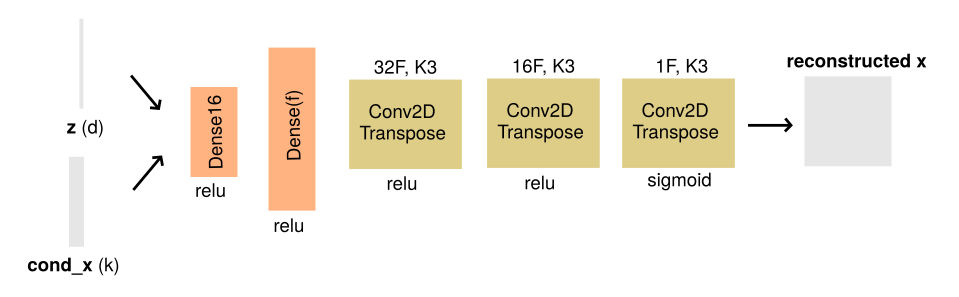}
        \caption{Decoder architecture}
        \label{fig:architecture-dec}
    \end{subfigure}
    \begin{subfigure}[t]{0.88\textwidth}
        \centering
        \includegraphics[width=\columnwidth]{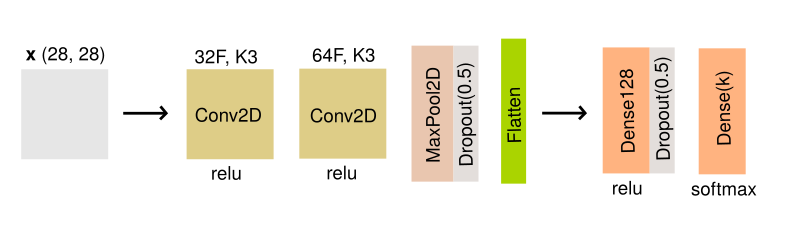}
        \caption{Classifier architecture}
        \label{fig:architecture-classifier}
    \end{subfigure}
\end{figure}

The encoder and the decoder parts of the CVAE architecture, and the classifier used are described in Figure \ref{fig:architecture-enc}, \ref{fig:architecture-dec} and \ref{fig:architecture-classifier} respectively. The latent dimension ($d$) is chosen per dataset. For MNIST, $d=8$ and for Fashion MNIST, $d=10$. The number of features after the convolutions in the encoder is represented by $f$. ``cond\_x'' is the one hot representation of class labels. $k$ in the classifier architecture represents the number of classes in the training data.

\section{Metrics Definitions}
The definitions of metrics used to evaluate OOD detection are as follows.

\textbf{FPR at 95\% TPR} is the probability of an OOD input being misclassified as in-distribution when 95\% of in-distribution samples are correctly classified as in-distribution (i.e, the true positive
rate (TPR) is at 95\%). True positive rate is calculated as, $TPR = \frac{TP}{TP + FN}$, where TP and FN denote the true positives and false negatives, respectively. The false positive rate (FPR) is computed as $FPR = \frac{FP}{FP+TN}$, where FP and TN denote the false positives and true negatives, respectively.

\textbf{Detection error} is the minimum mis-classification probability
over all possible thresholds over the OOD score. We assume that the test set contains equal number of in and out of distribution samples.

\textbf{AUROC} is the area under the receiver operating characteristic curve, which is a threshold independent metric. ROC curve is a plot of TPR versus FPR. AUROC can be interpreted as the probability that a positive example is assigned a higher detection score than a negative example. For a perfect detector, AUROC is 100\%.

\textbf{AUPR} is the Area under the Precision-Recall (PR) curve. PR curve is a plot of precision ($TP=(TP + FP)$) versus
recall ($TP=(TP + FN)$). The metric AUPR-In and AUPR-Out represent the area under the PR curve depending on if in or out of distribution data are specified as positives, respectively.

\section{More Results}

We report the OOD detection results for OOD detection methods based on softmax-score (\cite{hendrycks2016baseline}), uncertainty of classifier obtained via MC-dropout (\cite{gal2016dropout}), and mutual information between predictions and model posterior (\cite{Gal_bald_2017}). The results are shown in Table \ref{more results-table}.

\begin{table}
    \scriptsize
  \caption{Results}
  \label{more results-table}
  \centering
  \hskip-1.5cm\begin{tabular}{lllllll}
    \toprule

    \multirow{2}{*}{\makecell{ID Model}}     & \multirow{2}{*}{OOD}    & \makecell{FPR at \\ 95\% TPR \textdownarrow} & \makecell{Detection \\ Error\textdownarrow} & \makecell{AUROC\textuparrow} & \makecell{AUPR\\Out\textuparrow} & \makecell{AUPR\\In\textuparrow} \\
    \cmidrule(r){3-7}
    & & \multicolumn{5}{c}{Softmax/MC-Dropout/Mutual-Info} \\
    \midrule

    \multirow{7}{*}{\makecell{MNIST}} & F-MNIST  & 1.61/2.1/54.3 & 3.3/3.5/14.5 & 99.5/99.4/90.8 & 99.5/99.5/91.5 & 99.5/99.4/86.2 \\
    & EMNIST-letters & 28.0/24.5/22.0 & 12.6/11.2/11.1 & 93.6/94.6/95.0 & 93.4/94.4/94.7 & 93.0/94.1/94.7  \\
    & NotMNIST & 13.1/12.5/22.1 & 7.1/6.7/9.1 & 97.4/97.6/95.5 & 97.8/97.9/96.0 &  97.0/97.1/94.0
    \\
    & Omniglot & 0.0/0.0/92.8 & 0.5/0.6/19.5 & 100.0/100.0/84.5 & 100.0/100.0/88.7 & 100.0/100.0/73.9
    \\
    & Gaussian-Noise & 0.0/0.0/100.0 & 0.0/0.0/49.3 & 100.0/100.0/17.2 & 100.0/100.0/37.7 & 100.0/100.0/34.0
    \\
    & Uniform-Noise & 0.0/0.0/100.0 & 0.0/0.0/43.8 & 100.0/100.0/45.2 & 100.0/100.0/56.4 & 100.0/100.0/42.9
    \\
    & Sphere-OOD & 0.8/1.3/7.6 & 2.6/3.1/4.2 & 99.3/99.2/97.3 & 99.5/99.4/99.3 & 99.1/99.0/93.1\\
    \midrule

    \multirow{7}{*}{\makecell{F-MNIST}} & MNIST
    & 84.5/68.9/19.4  &  34.5/25.0/11.9 & 70.6/82.2/93.5 & 70.9/83.0/91.3 & 68.2/80.3/94.9   \\
    & EMNIST-letters & 88.1/77.8/37.6  & 41.1/33.1/21.1 & 62.3/73.4/84.2/ & 62.3/73.2/79.6 & 61.3/72.2/87.9 \\
    & NotMNIST & 83.1/67.0/24.2 & 35.2/25.1/14.3 & 68.9/81.7/91.7 & 66.4/80.3/87.6 & 68.6/80.8/93.6\\
    & Omniglot & 39.0/32.5/26.8 & 17.1/14.8/10.4 & 91.4/93.5/95.2 & 91.4/93.7/95.6 & 91.8/95.5/93.1
    \\
    & Gaussian-Noise & 99.1/98.5/72.2 & 17.5/15.8/9.6 & 80.0/82.1/92.5 & 87.8/89.1/95.3 & 65.5/67.9/83.4
    \\
    & Uniform-Noise & 96.9/96.5/48.8 & 29.4/24.2/16.8 & 70.1/76.8/90.9 & 78.6/84.0/92.5 & 58.11/64.8/87.9
    \\
    & Sphere-OOD & 97.2/71.2/1.8 & 50.0/17.0/2.5 & 48.4/88.2/99.6 & 50.6/91.4/99.5 & 47.7/82.4/99.6 \\
    \midrule

    \multirow{7}{*}{\makecell{MNIST0-4}} & MNIST5-9 & 18.2/15.7/15.3 & 10.6/9.7/9.9 & 94.2/94.8/94.5 & 93.0/93.1/92.8 &  94.8/95.4/95.1 \\
    & F-MNIST & 3.1/4.4/8.8 & 4.0/4.6/5.9 & 99.0/98.8.97.8 & 99.2/99.0/98.4 & 98.7/98.4/96.6 \\
    & EMNIST-LETTERS & 25.3/21.6/21.4 & 12.8/12.2/12.2 & 92.8/93.5/93.4 & 90.6/91.6/91.3 & 93.3/93.9/94.1   \\
    & NotMNIST & 21.4/16.2/15.9 & 10.44/9.3/9.4 & 95.5/96.4/96.5 & 95.5/96.4/96.1 & 95.0/95.9/96.4\\
    & Omniglot & 0.2/0.1/0.5 & 1.7/1.9/2.4 & 99.4/99.4/99.2 & 99.6/99.6/99.4 & 98.9/99.0/98.8\\
    & Gaussian-Noise &  0.0/0.0/0.0 & 0.3/0.4/1.0 & 99.7/99.7/98.8 & 99.8.99.8/99.4 & 98.9/98.9/95.8  \\
    & Uniform-Noise &   0.0/0.0/0.0 & 0.6/0.9/2.2 & 99.7/99.6/98.3 & 99.8/99.8/99.0 & 99.1/99.0/95.2 \\
    & Sphere-OOD &  1.0/1.9/2.5 & 2.8/3.4/3.6 & 99.2/99.0/98.6 & 99.4/99.3/99.1 & 98.6/98.5/97.5 \\
    \midrule
    \multirow{7}{*}{\makecell{F-MNIST0-4}} & F-MNIST5-9 & 73.5/67.1/32.8 & 26.1/23.4/17.3 & 80.1/83.4/89.8 & 80.4/83.3/87.6 & 78.1/82.0/91.2
    \\
    & MNIST &  44.9/43.9/73.9 & 15.9/16.0/16.9 & 91.0/91.4/87.7 & 91.2/91.3/89.5 & 90.4/90.5/82.3  \\
    & EMNIST-letters & 69.8/66.6/43.1 & 26.7/25.4/20.5 & 80.4/82.2/87.4 & 81.1/82.8/86.5 & 79.6/81.5/97.9 \\
    & NotMNIST & 71.9/67.0/38.6 & 24.3/20.8/17.0 & 82.8.86.1/90.8 & 84.3/87.7/90.4 & 80.2/83.4/91.0 
    \\
    & Omniglot & 44.8/41.3/29.6 & 15.3/14.0/11.5 & 91.6/93.0/94.7 & 92.1/93.8/94.9 & 91.0/92.2/93.9
    \\
    & Gaussian-Noise & 0.3/0.5/98.3 & 2.3/2.3/12.2 & 99.8/99.7/87.3 & 99.8/99.8.92.4 & 99.7/99.7.74.2
    \\
    & Uniform-Noise & 27.7/20.5/8.4 & 8.6/7.9/5.3 & 96.4/97.2/98.0 & 97.2/97.8/98.6 & 95.5/96.5/96.6 
    \\
    & Sphere-OOD & 86.5/67.3/10.7 & 33.7/20.4/7.8 & 71.6/86.2/97.3 & 73.4/88.0/96.7 & 67.3/83.0/97.8\\

    \bottomrule
  \end{tabular}
\end{table}

\end{document}